\pdfoutput=1
\documentclass[11pt]{article}

\usepackage[final]{acl}

\usepackage{times}
\usepackage{latexsym}

\usepackage[T1]{fontenc}
\usepackage[utf8]{inputenc}

\usepackage{microtype}

\usepackage{inconsolata}

\usepackage{graphicx}
\usepackage{booktabs}
\usepackage{multirow}
\usepackage{amssymb}
\usepackage{amsmath}
\usepackage{cuted}
\usepackage{multicol}
\usepackage{hyperref}
\usepackage{url}
\usepackage{tcolorbox}
\usepackage{tabularray}
\usepackage{soul, color, xcolor}
\usepackage{makecell}
\usepackage{algorithm}
\usepackage{algorithmic}
\usepackage{color, colortbl}

\definecolor{lightgoldenrodyellow}{rgb}{0.98, 0.98, 0.82}

\title{KSOD: Knowledge Supplement for LLMs On Demand}

\author{Haoran Li, Junfeng Hu, \\
 National Key Laboratory for Multimedia Information Processing, \\
    School of Computer Science, Peking University \\
\texttt{haoranli@stu.pku.edu.cn} \\
\texttt{hujf@pku.edu.cn}
}

\begin{document}
\maketitle

\begin{abstract}

Large Language Models (LLMs) have demonstrated remarkable capabilities in various tasks, yet still produce errors in domain-specific tasks. 
To further improve their performance, we propose KSOD (Knowledge Supplement for LLMs On Demand), a novel framework that empowers LLMs to improve their capabilities with knowledge-based supervised fine-tuning (SFT). 
KSOD analyzes the causes of errors from the perspective of knowledge deficiency by identifying potential missing knowledge in LLM that may lead to the errors. Subsequently, KSOD tunes a knowledge module on knowledge dataset and verifies whether the LLM lacks the identified knowledge based on it. If the knowledge is verified, KSOD supplements the LLM with the identified knowledge using the knowledge module.
Tuning LLMs on specific knowledge instead of specific task decouples task and knowledge and our experiments on two domain-specific benchmarks and four general benchmarks empirically demonstrate that KSOD enhances the performance of LLMs on tasks requiring the supplemented knowledge while preserving their performance on other tasks.
Our findings shed light on the potential of improving the capabilities of LLMs with knowledge-based SFT.

\end{abstract}

\section{Introduction}

Large Language Models (LLMs) have demonstrated excellent performance across a wide range of tasks, showing their remarkable general-purpose capabilities\citep{openai-gpt3,ouyang2022training,gpt4,touvron2023llama2,chowdhery2023palm,jiang2024mixtral}. However, LLMs still hallucinate and produce 
factually incorrect, irrelevant, or incomplete content, leading to the errors in their outputs.

\begin{figure}[t]
  \includegraphics[width=\columnwidth]{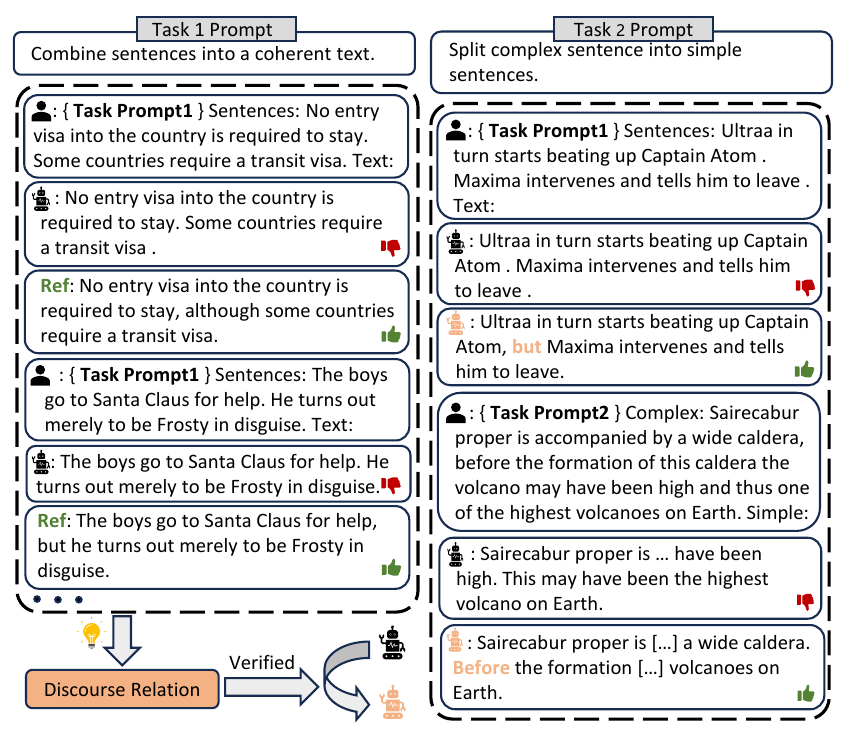}
  \caption{On the left side of the figure, samples from Task 1 is presented in the form of (input, output with errors, correct reference). Based on these samples, KSOD identifies the missing knowledge as discourse relations. After verify that LLM lacks this knowledge, it is supplemented into the LLM. As shown on the right side of the figure, after supplementation, the model generates correct outputs not only for Task 1 but also for another task (Task 2) that requires the discourse relation knowledge.}
  \label{fig:teaser}
\end{figure}

\begin{figure*}[t]
  \centering  
  \vspace*{-5pt}
  \includegraphics[width=\textwidth]{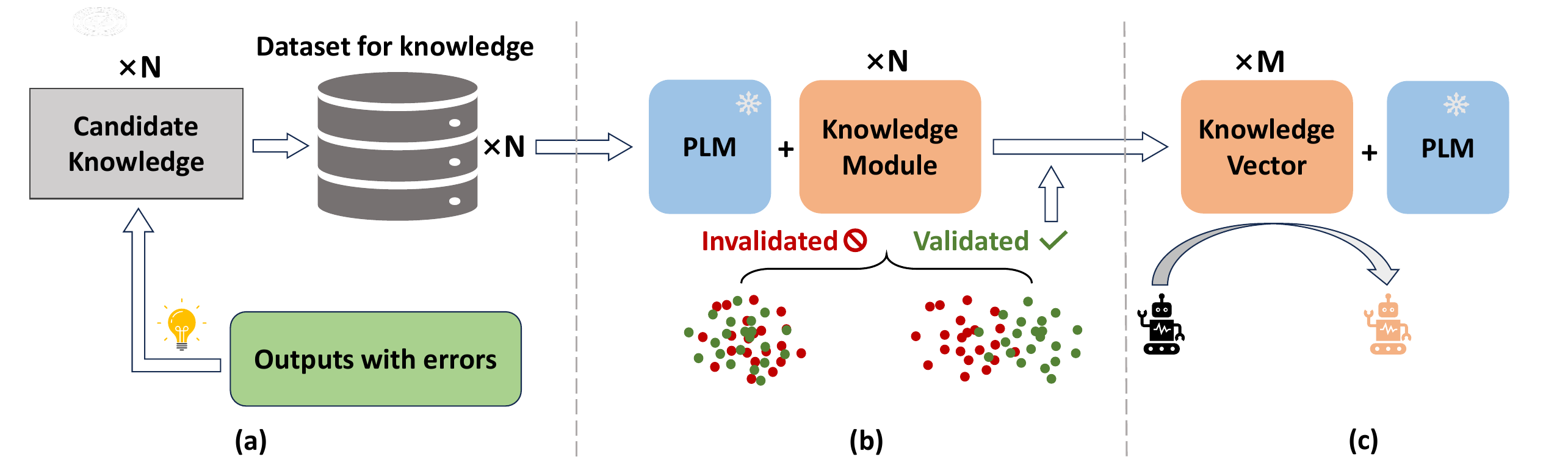}
  % \vspace*{-5pt}
  \caption{Our KSOD framework consists of three stages: (a) Knowledge Identification; (b) Knowledge Verification; (c) Knowledge Supplement.}
  \label{fig:overview}
  % \vspace*{-10pt}
  \vspace*{-5pt}
\end{figure*}

Existing methods\citep{an2023learning, ying-etal-2024-llms, tong2024can} for improving the outputs of LLMs commonly depend on supervised fine-tuning (SFT). These methods necessitate extensive training datasets generated with stronger LLMs (e.g. GPT-4) or costly human annotations, which may not always be accessible. Furthermore, the application of SFT on datasets collected from some tasks can potentially compromise the capabilities of LLMs on other tasks. Consequently, many studies explore the potential of self-correction, where the LLM itself is prompted or guided to repair the errors in its own output by refining the output\citep{pan2023automatically}. 
Despite the self-correction method improves the fluency and understandability, it leads to false positive optimization and reduces diversity in text generation because of the universal existence of self-bias\citep{xu2024pride}. Moreover, LLMs cannot solve the errors caused by the lack of knowledge or ability based on the feedback generated by themselves.

To address these challenges, we introduce the KSOD (Knowledge Supplement for LLMs On Demand) framework to correct the errors by supplementing LLMs with required knowledge on demand. 
KSOD framework diverges from conventional SFT methods by decoupling knowledge and task to correct the errors of specific task from the perspective of knowledge rather than the task itself.
As illustrated in Figure \ref{fig:teaser}, supplementing LLMs with missing knowledge, identified as the cause of errors in Task 1, not only mitigates these errors within in task 1, but also yields consistent improvements in tasks that also rely on the same knowledge, such as Task 2. Furthermore, our empirical evaluation on general tasks that explicitly require this knowledge demonstrates that KSOD introduces little to no degradation to LLMs' performance in these tasks.

In specific, KSOD first identifies the knowledge missing in LLMs that may lead to the errors and collects dataset containing the required knowledge from existing datasets. Subsequently, we train a knowledge module on the identified dataset. 
To ensure that the learned knowledge is genuinely missing in LLMs rather than already possessed, which could introduce noise, KSOD performs a knowledge verification step. Notably, only knowledge modules that pass this verification phase proceed to the knowledge supplement stage, where they are injected into the LLM.
Finally, the verified knowledge module is integrated into LLMs to correct errors arising from missing knowledge. Figure \ref{fig:overview} provides an overview of our framework.

To validate the performance of KSOD, we conduct comprehensive experiments with open source LLMs on both two error-prone tasks where the errors occurred and four general tasks. Our results show that supplementing knowledge with KSOD brings a notable reduction in errors, leading to notable performance improvements. Furthermore, after supplementing the missing knowledge through KSOD, the performance of LLMs on four general tasks and the remaining error-prone task remains unchanged or slightly decreases, with some cases even showing improvement. These empirical findings demonstrate that KSOD effectively correct errors by supplementing the desired knowledge missing in LLMs while preserving the LLMs' performance on other tasks. This finding highlights the potential of enhancing LLMs performance by supplementing knowledge in a task-agnostic way.

In summary, our contributions are as follows:
\begin{itemize}
    \item Given specific knowledge, KSOD provides a general method to verify whether LLM lacks the knowledge. 
    % To the best of my knowledge, this is the first approach that explores how to assess whether LLMs lack specific knowledge.
    \item We propose the KSOD framework, which corrects the errors from LLMs by supplementing required knowledge on demand with knowledge-based SFT, while preserving LLM's performance on other tasks. 
    % We propose a simple but effective knowledge supplement method within the KSOD framework to supplement verified knowledge. 
    % Editing the model from the perspective of knowledge rather than tasks makes model editing simpler, enhances generalization capabilities, and improves interpretability.
    \item With extensive experiments, we validate the effectiveness of our proposed framework and reveal the potential of enhancing LLMs' performance by supplementing knowledge in a task-agnostic way.
    % Experimental results demonstrate that it is feasible to enhance the task-specific capabilities of LLMs without harming the performance of LLMs on other tasks through supplementing required knowledge absent in LLMs. 
    % This finding revealing the potential of separating knowledge from pre-trained parameters and allowing LLMs to acquire knowledge in a ’verify-and-supplement‘ manner.
\end{itemize}

\section{KSOD Framework}
\subsection{Proposed Research Questions}
To correcting the errors in outputs of LLMs from the perspective of knowledge, we try to research knowledge-based SFT to supplement the knowledge desired to generate correct outputs but missing in LLMs. Specifically, we aim to address the following research questions (RQs):
\begin{itemize}
    \item \textbf{RQ1}: How to verify whether LLMs lack specific knowledge?
    \item \textbf{RQ2}: What are the effects of knowledge-based SFT on tasks that require such knowledge and those that do not?
\end{itemize}

\subsection{Preliminaries: Knowledge-based SFT}
\label{part:scope}
We begin with preliminaries, formally introducing knowledge-based SFT and the scope of two RQs.

% \textbf{Scope of three RQs.} 
% \textbf{Objective of Knowledge-based SFT.} 
In the context of LLMs, the knowledge can be implicit knowledge encoded within the model's parameters. Given a LLM represented by parameter $\theta_0$, the objective of knowledge-based SFT is update the parameters of LLM as $\theta' = \theta_0 + \Delta \theta$, where $\Delta \theta$ is a low-rank update relative to $\theta_0$ and encodes a specific type of knowledge that is missing in the original LLM. 

In this study, we focus on knowledge of categories that can be formalized as a classification task. Therefore, the scope of two RQs are restricted to knowledge learned as a classification task in this work.
% To supplement the lacking knowledge without damaging the general capabilities of LLMs, the $\Delta \theta$ should be low-rank compared to $\theta_0$. 

\subsection{KSOD Overview}
In this section, we present our knowledge-based SFT framework, KSOD, to correct the errors in outputs of LLMs from the perspective of knowledge. 
As shown in Figure \ref{fig:overview}, KSOD consists of three stages: Knowledge Identification (\S\ref{part:one}), Knowledge Verification (\S\ref{part:two}) and Knowledge Supplement (\S\ref{part:three}). 

The knowledge identification stage aims to find dataset containing the missing knowledge, whose deficiency may cause the errors in outputs of LLMs. During knowledge verification stage, KSOD fine-tunes the LLMs on these datasets using LoRA\citep{hu2021lora} as a knowledge module and verifies whether the LLMs lack specific knowledge based on the embeddings distribution of the knowledge module. Clearly, not all knowledge identified in the knowledge identification stage is missing in the LLM. Only the knowledge that passes verification will be passed to the knowledge supplement stage. During this stage, the verified knowledge module will be supplemented into the LLMs to enhance the performance of LLMs on the tasks that requiring the knowledge.

\section{Verifying whether the LLM lacks specific knowledge}

This section focus on the verification of identified knowledge and aims to solve RQ1. The verification process is outlined in Algorithm \ref{alg:KI_KC}. 

\begin{algorithm}
\caption{Process of knowledge identification and knowledge verification stages.}
\label{alg:KI_KC}
\setlength{\multicolsep}{0.0pt plus 0.0pt minus 0pt}
    \begin{algorithmic}[1]
        \REQUIRE{Language model $\pi_{\theta_0}$, a set of samples $\mathcal{F}=\{s_1, ..., s_K\}$, expert model $E$, classification layer $\pi_{\theta_c}$, knowledge module $\pi_{\Delta \theta}$, threshold $\epsilon$}
        \ENSURE{A set of verified knowledge module $\mathcal{M}$}
        \STATE $\mathcal{K} \gets E(\mathcal{F})$ 
        \STATE $\mathcal{M} \gets \{\}$
        \FOR{knowledge $k$ \textbf{in} $\mathcal{K}$}
        \STATE Dataset $d_k \gets$ Collect(k)
        \STATE Update $\pi_{\theta_c}$ with $\pi_{\theta_0}$ frozen on $d_k$
        \STATE Update $\pi_{\Delta \theta}$ with $\pi_{\theta_0}$ and $\pi_{\theta_c}$ frozen on $d_k$
        \STATE $E_k \gets Embed(\pi_{\Delta \theta}, d_k)$
        \STATE $S_k \gets$ S\_C$(E_k, d_k)$
        \IF{$S_k \geq \epsilon$}
        \STATE $\mathcal{M} \gets \{ \pi_{\Delta \theta} \} \cup \mathcal{M}$
        \ENDIF
        \ENDFOR
    \end{algorithmic}
\end{algorithm}

This algorithm corresponds to stages (a) and (b) in Figure \ref{fig:overview}. Specifically, Collect($\cdot$) refers to gathering the dataset corresponding to the knowledge (line 4 of Algorithm \ref{alg:KI_KC}), Embed($\cdot$) denotes obtaining the distribution of embeddings from knowledge module on $d_k$ (line 7 of Algorithm \ref{alg:KI_KC}) in and S\_C($\cdot$) represents the computation of the Silhouette Coefficient (line 8 of Algorithm \ref{alg:KI_KC}).

\subsection{Knowledge Identification}
\label{part:one}

% A task $F_t$ usually appear in the form of dataset $\{D_{t,i}|i=1,2,...,n\}$ where LLMs need to complete the mapping from input $x$ to output $y$ on the test dataset. 
To learn from errors from the perspective of knowledge, KSOD identifies the knowledge whose deficiency in LLMs may cause the errors. 

Formally, given a set $\mathcal{F}$ consisting of samples in the format of (input, erroneous output, correct reference), the aim of knowledge identification stage is to construct the set that contains the knowledge whose absence in LLMs may cause the errors in $\mathcal{F}$. This set can be denoted as $\mathcal{K} = \{k_{1}, k_{2}...\}$. 

To construct $\mathcal{K}$, we manually select $N$ samples with similar errors from $\mathcal{F}$. Leveraging the powerful knowledge storage and language processing capabalities, we utilize strong LLMs like GPT-4 to identify the knowledge whose absence in LLM may cause the errors in $\mathcal{F}$. The simplified parompt template is as follows:

\begin{tcolorbox}[colback=gray!5!white,colframe=gray!75!black,title=Prompt for Knowledge Identification]
    \{TASK DEFINITION\} Please analyze the errors that arise in output of \{TASK NAME\} task in the given samples.  \\
    \{ \\
    $\quad$  sample $i$:  \\
    $\quad$  Input: \{input text\} \\
    $\quad$  Target: \{correct reference\} \\
    $\quad$  Output: \{output with errors\} \\
    % \} $\times$ $N$ samples
    \} $i$ = 1,2,...,$N$ \\
    Firstly, provide a step-by-step analysis for the common characteristics of the errors from all samples. \\
    Next, identify the potential knowledge lacking in LLM that may have led to these errors.
\end{tcolorbox}

After obtaining $k_{i}$, the process of finding the datasets containing knowledge $K_{i}$ from available existing NLP datasets becomes more straightforward and simple. For sample, huggingface\footnote{https://huggingface.co/datasets} offers more than 250K datasets where we can search and download dataset containing the identified knowledge. 

\begin{table*}[]
\tiny
\centering

% \scriptsize
\resizebox{0.95\textwidth}{!}{%
\begin{tabular}{
@{}
c@{\hspace{1.3ex}}
% p{2cm}@{\hspace{2ex}}
p{3.7cm}@{\hspace{2ex}}
% p{3.7cm}@{\hspace{2ex}}
% p{3.2cm}
p{3.7cm}
@{}
}
\toprule

% \multirow{1}{*}{\rotatebox[origin=c]{90}{\bf Factuality }}  
\multirow{-1.4}{*}{\rotatebox[origin=c]{90}{\bf Examples with errors }}  

  % \multirow{1}{*}{\rotatebox[origin=c]{90}{\bf Completeness}} 
  &  
  % \emph{(human)} 
  % \textbf{Seems unsettled but here’s a theory}: The most recent hypothesis suggests that the hair loss process begins during puberty, when growth of the skull and the muscles in the forehead and neck increases the tension in a tight band of tissue stretching over the top of the head. The more DHT (a type of testosterone) there is... 
  \textbf{Output}: He [...] managed eventually to pay his creditors in full. Some inheritance was left for his descendants.  &
%   , the more these muscles contract, which causes inflammation and brings in more DHT. [...] &
  % \emph{(expert)} It is difficult to decide but I believe 
  % \emph{(expert)} \hl{Answer A is the better choice as it describes both the hormonal and environmental causes and Answer B only focuses on one theory which might not be 100 percent accurate.} [...] According to research, baldness is due to genes. In 95 percent cases, balding is due to androgenetic alopecia [...] %, more commonly known as male pattern baldness, which is hereditary condition.  
  \textbf{Target}: He [...] managed eventually to pay his creditors in full \hl{so that} some inheritance was left for his descendants.
   \\

   \cmidrule{2-3}
  
  % \multirow{1}{*}{\rotatebox[origin=c]{90}{\bf Completeness}} 
  
  &  
  
  % \emph{(human)} 
  % \textbf{Seems unsettled but here’s a theory}: The most recent hypothesis suggests that the hair loss process begins during puberty, when growth of the skull and the muscles in the forehead and neck increases the tension in a tight band of tissue stretching over the top of the head. The more DHT (a type of testosterone) there is... 
  \textbf{Output}: They finished third among the league's eight teams, with Gore as their starting center fielder. O'Rourke had moved to left due to the departure of Slattery.  &
%   , the more these muscles contract, which causes inflammation and brings in more DHT. [...] &
  % \emph{(expert)} It is difficult to decide but I believe 
  % \emph{(expert)} \hl{Answer A is the better choice as it describes both the hormonal and environmental causes and Answer B only focuses on one theory which might not be 100 percent accurate.} [...] According to research, baldness is due to genes. In 95 percent cases, balding is due to androgenetic alopecia [...] %, more commonly known as male pattern baldness, which is hereditary condition.  

  \textbf{Target}: They finished third among the league's eight teams, with Gore as their starting center fielder, \hl{while} O'Rourke had moved to left due the departure of Slattery. \\

\midrule

\multirow{-1.2}{*}{\rotatebox[origin=c]{90}{\bf Expert Justification }}

    &  \textbf{Knowledge Type}: Understanding of Logical and Causal Relationships. \emph{(GPT-4o)}  &
%   , the more these muscles contract, which causes inflammation and brings in more DHT. [...] &
  % \emph{(expert)} It is difficult to decide but I believe 
  The model fails to detect and explicitly represent causal or logical links [...] \emph{(GPT-4o)} 
   \\
   
  % \midrule
  \cmidrule{2-3}
  
  % \multirow{1}{*}{\rotatebox[origin=c]{90}{\bf Completeness}} 
  &
  
%   settlement on how much blood flow there is to an area and if there's any muscle action to help break them up. [...] 
 \textbf{Knowledge Type}: Discourse Structure Understanding. \emph{(DeepSeek-R1)}  &
%   For example, clots in the veins are usually broken down more slowly than clots in the arteries. &
  LLMs struggle to recognize implicit discourse relationships (e.g., cause-effect, contrast) between sentences and select appropriate connectives (as a result, while, so that). [...] \emph{(DeepSeek-R1)} \\

  \bottomrule
\end{tabular}
}
\vspace{-0.6em}
\caption{One example of knowledge identification with two strong LLMs. This table presents 2 out of the 4 examples; the full prompt can be found in Appendix \ref{app:identification}. }
\label{tab:KI_Results}
\vspace{-0.15in}
\end{table*}

\begin{figure*}[t]
  \includegraphics[width=\textwidth]{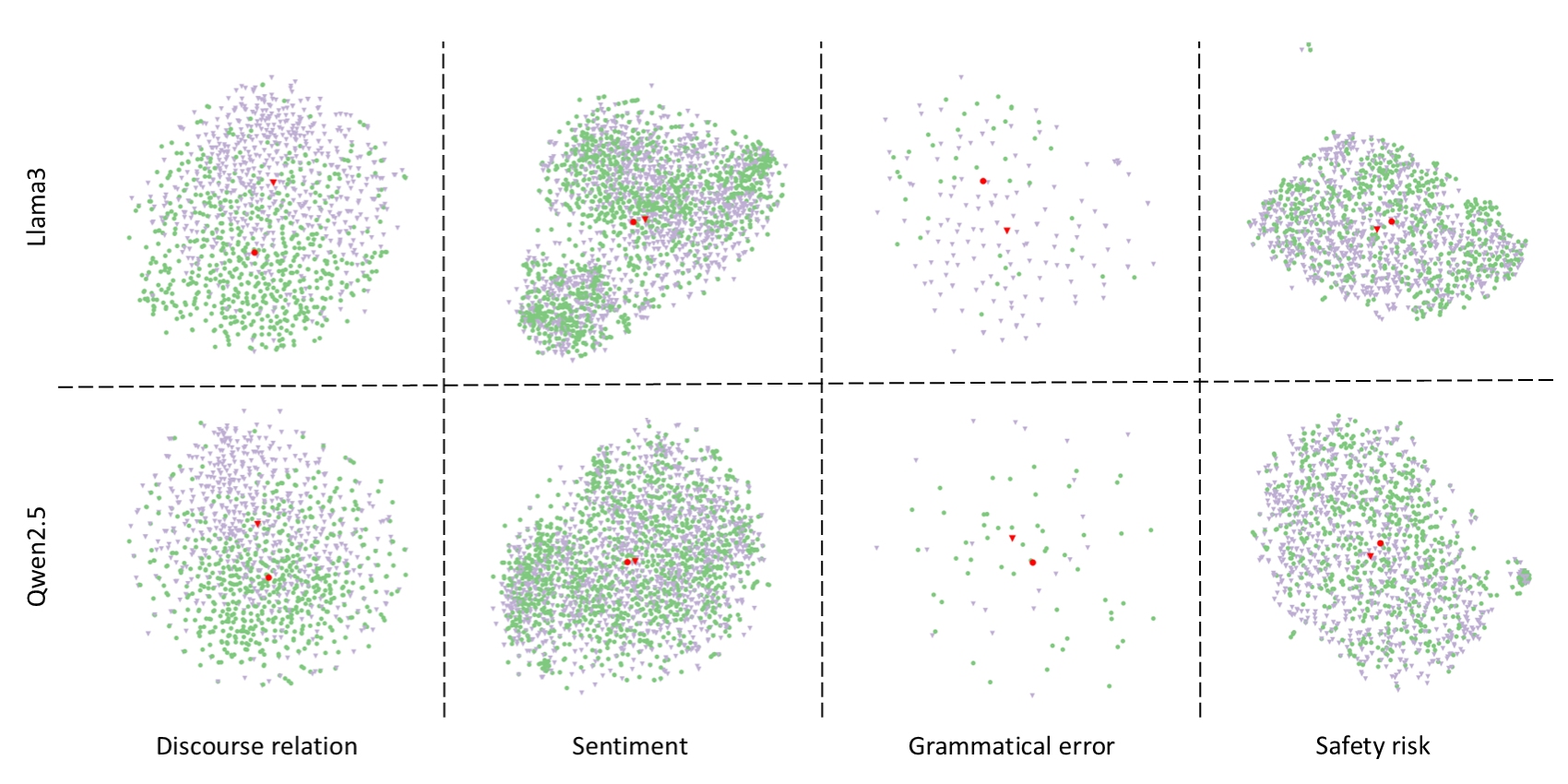}
  \caption { T-SNE \citep{van2008visualizing} visualization of the embedding distribution and each color represents a category within categorical knowledge based on dataset labels. The embedding is the last token embedding from B matrix of LoRA on test set.}
  \label{fig:visualization}
\end{figure*}

\subsection{Knowledge Verification}
\label{part:two}
After fine-tuning a LLM on the target task using LoRA, its performance in maintaining capabilities on other tasks surpasses full fine-tuning, and even common regularization methods \citep{biderman2024lora}. Therefore, LoRA is a suitable method for LLMs to learn the knowledge they lack without affecting the initial capabilities of LLMs on other tasks. Based on the hypothesis that the change in weights during fine-tuning is low rank, the vanilla LoRA is mathematically represented as:
% \begin{align}
% W’ = W_0 + \Delta W = W_0 + \frac{\alpha}{r}BA = W_0 + \eta BA
% \label{eq:importance_score}
% \end{align}
\begin{equation}
	\begin{split}
        W’  &= W_0 + \Delta W \\
            &= W_0 + \frac{\alpha}{r}BA \\
            &= W_0 + \eta BA
	\end{split}
        \label{eq:importance_score}
\end{equation}

where $W', W_0 \in \mathbb{R}^{m\times n}$, $B\in \mathbb{R}^{m\times r}$, and $A\in \mathbb{R}^{r\times n}$, with $r \ll \min(m,n)$. $W_0$ is the pre-trained weight matrix and $\eta$ is a hyperparameter serving as a scalar weight, where both of them are frozen during fine-tuning. Only $A$ and $B$ contain trainable parameters. As stated in Section \ref{part:scope}, knowledge-based SFT in this work is performed in the form of a classification task. Therefore, we additionally introduce a classifier layer, which is also trainable.

However, when utilizing the LoRA to learn knowledge, ideally, the scalar of LoRA should be large when current knowledge is deficient and employ the parameters in $A$, $B$ for learning. Conversely, the scalar should be small when the knowledge is sufficient, thus avoid introducing noise. To enable LoRA adeptly adjust the scalar value based on different knowledge, we follows \citet{liu2024sparsely} and set $\eta$ to be trainable, which has already been proven to be an important design for improving LoRA's performance. 
% Moreover, a higher learning rate is allocated to $\eta$, allowing LoRA to quickly adjust scalar based on the usefulness of the knowledge learned by $A$ and $B$. 

Furthermore, to further reduce the impact of LoRA on the general capabilities of LLMs, we divided the training of LoRA on knowledge dataset into two stages: in the first stage, only the classification layer is tuned with LLM frozen; in the second stage, only LoRA is tuned with both LLM and the tuned classification layer frozen. 

Based on the LoRA with trainable scalar, we call the LoRA variant, which is fine-tuned on a specific knowledge dataset, a knowledge module. We hypothesize that the embedding distribution of knowledge module will exhibit clustering characteristics consistent with knowledge categorization if and only if the LLMs lack the knowledge. To prove this hypothesis, we examine the embedding distribution of knowledge modules tuned on different knowledge datasets in Section \ref{part:exp_kv}.  

In summary, the Knowledge Verification stage can be divided into two steps as shown in Algorithm \ref{alg:KI_KC}: initially, we fine-tune LoRA based on the dataset procured during the knowledge identification stage to obtain the knowledge module; subsequently, we evaluate the effectiveness of the knowledge module by verifying whether the embedding distribution of the knowledge module exhibits clustering characteristics consistent with knowledge categories in dataset. If these clustering characteristics do not become apparent, it is inferred that the LLM does not lack this particular type of knowledge. 
As a result, we verify the next kind of knowledge identified during knowledge identification stage. If the clustering characteristics is apparent, the knowledge is verified and will be supplemented to LLMs during knowledge supplement stage. 
% As a result, we recycle the procedure of Knowledge Identification and Verification until an effective knowledge module has been developed.

\subsection{Experiment for knowledge identification}
\paragraph{Tasks and Datasets.}
We collect samples with errors in Sentence Fusion task, whose target is joining several independent sentences into a single coherent text. 
Specifically, we used DiscoFuse\citep{geva2019discofuse}, a large-scale sentence fusion dataset, to collect the initial outputs with errors from LLMs.

\paragraph{Experimental Setup.}
In terms of selecting LLMs for the experiments, we need to select two kinds of LLMs: the LLMs generate outputs with errors and the strong LLMs for identifying the missing knowledge that may cause the errors in outputs.
For the former, we employ 2 different open-source LLMs for experiments as follows:
\begin{itemize}
    \item \textbf{LLaMA-3.1-8B}~\cite{dubey2024llama}~(denoted as LLaMA-3) is a dense LLM with massive pre-training on extremely large corpora, which is developed by Meta.
    \item \textbf{Qwen2.5-7B}~\cite{yang2024qwen2}~(denoted as Qwen2) is a powerful multilingual LLM developed by Alibaba Cloud.
\end{itemize}
For the latter, we employ GPT-4o \citep{hurst2024gpt} and DeepSeek-R1 \citep{guo2025deepseek} to identify knowledge.

We manually selected 4 samples with similar errors and used the prompt presented in Section \ref{part:one} to analyze the potentially missing knowledge. 

\paragraph{Identified Knowledge.} 
As shown in Table \ref{tab:KI_Results}, all of the samples with errors have difficulty in utilizing proper conjunctions. Consequently, both GPT-4o and DeepSeek-R1 concluded that discourse relations constitute the most probable knowledge whose absence causes the errors in samples.
Therefore, we select discourse relation as the first knowledge to be verified. 

\subsection{Experiment for knowledge verification}
\label{part:exp_kv}

\begin{table}
  \centering
  \resizebox{1\columnwidth}!{\begin{tabular}{llllll}
    \hline
    \textbf{Dataset}     & \#\textbf{Class}      & \#\textbf{Train} & \#\textbf{Dev} & \#\textbf{Test} & \textbf{Deficiency} \\
    \hline
    DiscoWiki  &4   & 20,000     & 2,500       & 2,500   & Yes       \\
    SST-2      &2   & 20,000     & 2,500       & 2,500   & No       \\
    EXPECT     &15  & 15,187     & 2,413       & 2,416   & Yes       \\
    AEGIS2.0   &2   & 21,446     & 1,087       & 1,567   & No       \\
    \hline
  \end{tabular}}
  \caption{\label{tab:dataset}
    Data Statistics of four datasets for knowledge verification. 
    For datasets with a large size (DiscoWiki, SST-2), we sample the same number of instances from the dataset itself as the experimental dataset. 
  }
\end{table}

\paragraph{Tasks and Datasets.}
To get the corresponding dataset for discourse relation classification, we use an automatically rule-based method \citep{ma-etal-2019-implicit} to label the WikiSplit++ \citep{tsukagoshi2024wikisplit++} dataset, obtaining a dataset with discourse relation labels, which we called DiscoWiki. DiscoWiki contains four types of discourse relation following PDTB3.0\citep{webber2019penn} and we have selected an equal number of samples for each type of discourse relations.
To validate the effectiveness of the knowledge verification method, we introduce three additional types of knowledge and determine whether the LLM lacks them based on existing research:
\begin{itemize}
    \item \textbf{Discourse relation:} DiscoWiki is used for 4-class classification of discourse relations. According to the evaluation by \citet{chan-etal-2024-exploring}, LLMs still struggle to classify implicit discourse relations. Therefore, we conclude that this knowledge is missing in the LLM.
    \item \textbf{Sentiment:} Stanford Sentiment Treebank binary (SST-2) is used for 2-class sentiment classification. The binary sentiment classification is typically well mastered by LLMs and LLMs do not lack it.
    \item \textbf{Grammatical error:} EXPECT \citep{fei2023enhancing} is used for 15-class classification of grammatical error types. The error types can be used to enhance the performance of LLMs on Grammatical Error Correction (GEC) task\citep{fei2023enhancing}, where LLMs often underperform task-specific models in this task\citep{davis2024prompting}. The LLMs lack this knowledge so that supplementing LLMs with it allows for performance improvement on the GEC task.
    \item \textbf{Safety risk:} AEGIS2.0 \citep{ghosh2024aegis2} is used for 2-class classification of safety risks. \citet{zheng2024prompt} have found that LLMs are naturally capable of distinguishing harmful and harmless queries without safety prompts. Hence, the safety risks knowledge is not missing in LLMs. 
\end{itemize}

The detailed statistics of four datasets have listed in Table \ref{tab:dataset}.

\paragraph{Experimental Setup.} To learn knowledge with the selected dataset, we first tune the final classification linear layer itself with backbone parameters frozen. After tuning the classification linear layer, we learning knowledge with LoRA. Specifically, we set the init scalar $\eta$ for LoRA to 0, and both A and B are initialized with Gaussian initialization. More details of training can be found in Appendix \ref{app:verification}.

\begin{table}
  \centering
  \begin{tabular}{lll}
    \hline
    \textbf{Dataset}           & \textbf{Llama3} & \textbf{Qwen2}  \\
    \hline
    DiscoWiki      & 0.0423           & 0.0233             \\
    SST-2          & 0.0098           & 0.0027             \\
    EXCEPT         & 0.0478           & 0.0663             \\
    AEGIS2.0       & 0.0125           & 0.0108             \\
    \hline
  \end{tabular}
  \caption{\label{tab:KV_results}
    SC for embeddings clustering. 
  }
\end{table}

\begin{table*}[t]
% \resizebox{1.\linewidth}{!}
% {
% \scriptsize
\begin{tabular}{lccccc|c|c}
\toprule
\textbf{Model} & \multicolumn{5}{c|}{\textbf{General Language Tasks}} & \multicolumn{1}{c|}{\textbf{Sentence Fuse}} & \textbf{GEC}  \\
& Drop & Squad & ARC & HellaSwag &Avg. & DiscoFuse & CoNLL14 \\
\midrule
% \rowcolor{lightgoldenrodyellow}
LLaMA3-8B & 47.25 & 71.81 & 79.44 & 79.52 &69.51 & 43.89 & 37.14  \\
% LLaMA3-8B-DR & 47.28 & 72.01 & 79.27 & 79.65 &69.55 & 44.57 & 36.72 \\
LLaMA3-8B-DR & 47.16 & 71.74 & 79.35 & 79.66 &69.48 & 45.13 & 36.49 \\
LLaMA3-8B-GE & 46.88 & 71.62 & 79.61 & 79.70 &69.45 & 44.65 & 37.74 \\
% LLaMA3-8B-DR+GE & 46.52 & 71.45 & 79.52 & 79.76 &69.31 & 45.11 & 35.69 \\
LLaMA3-8B-DR+GE & 46.13 & 71.24 & 79.35 & 79.80 &69.13 & 45.31 & 36.84 \\
\midrule
Qwen2-7B & 37.93  & 56.98 & 89.85 & 77.84 & 65.65  & 43.41 & 31.03  \\
Qwen2-7B-DR & 40.27  & 57.92 & 90.02 & 77.91 & 66.53  & 43.81 & 30.97  \\
Qwen2-7B-GE & 38.95  & 57.63 & 89.76 & 78.08 & 66.11  & 44.04 & 31.34  \\
Qwen2-7B-DR+GE & 41.31  & 58.61 & 89.85 & 78.21  & 67.00 & 44.07 & 30.58   \\
\bottomrule
\end{tabular}
% }
\caption{Comparison of evaluation results of knowledge supplement among several benchmarks. DR refers to discourse relation and GE refers to grammatical error.}
\label{tab:llm-comparison}
\end{table*}

\subsection{RQ1: How to verify whether LLMs lack specific knowledge?}

% In order to ensure comparability among different categories of knowledge,  data from more than two classes in the optimal way and finally combined them into 2 classes for comparison. 
To ensure the comparability of categorical knowledge with different numbers of types, we select two types of data with the most distinct embedding distribution as representatives of this categorical knowledge.
To assessing the clustering characteristics, we visualize the embeddings of LoRA trained on different datasets and models in Figure \ref{fig:visualization} and calculate the Silhouette Coefficient (SC) score with knowledge category label to evaluate the cluster characteristics in Table \ref{tab:KV_results}. 

Based the visualization of embeddings distribution of knowledge for four datasets, it it obvious that the embeddings distribution of discourse relation and grammatical error exhibits characteristics corresponding to knowledge categories. From the SC calculation results in the Table \ref{tab:KV_results}, we can also reach the same conclusion that the embeddings distribution of knowledge module learned on DiscoWiki and EXCEPT exhibits clustering characteristics matching knowledge categories, while the knowledge modules learned on a dataset like SST-2 that contains knowledge already mastered by LLMs do not exhibit such characteristics. 
% The experimental results preliminarily demonstrate the validity of our hypothesis.
The experimental results empirically validate the effectiveness of our knowledge verification method.

% \subsection{Knowledge supplement}
\section{Effect of knowledge-based SFT}
\label{part:three}
\subsection{Knowledge supplement}
A \textit{task vector} is built by the difference between the weights of a pre-trained model and the weights of the same model after fine-tuning on a task which specifies the direction and stride of fine-tuning. More importantly, simple arithmetic on \textit{task vector} can be used to control the behavior of the resulting model\citep{ilharco2022editing}. Inspired by \textit{task vector}, we propose \textit{knowledge vector}, which can be built simply use the weights of knowledge module that has been verified during knowledge verification stage. In this way, LLMs can learn specific knowledge through the addition of the corresponding \textit{knowledge vector}.

Compared with \textit{task vector}, \textit{knowledge vector} decouples task and knowledge. The \textit{task vector} learns knowledge of a specific task, simultaneously, influences the original task instructions following ability to utilize corresponding task-specific knowledge, leading to the performance declines on other tasks \citep{kotha2023understanding,jiang2024interpretable,sun2024reviving}. Conversely, the \textit{knowledge vector} learns knowledge in a task-agnostic manner, exerting less impact on the general capabilities of LLMs.

\subsection{Experiment for knowledge supplement}

\paragraph{Tasks and Datasets.}
The evaluation is performed on four key general benchmarks using the LLMBox \citep{tang2024llmbox}, a comprehensive library for implementing LLMs, including a unified training pipeline and comprehensive model evaluation. 
We evaluate the LLM with knowledge vector with four benchmarks for general language tasks. Furthermore, we incorporate benchmarks requiring the verified knowledge, including an augmented version of DiscoFuse with multi-reference\citep{ben2020semantically} for discourse relation knowledge and CoNLL14 \citep{ng2014conll} for grammatical error knowledge.

\subsection{RQ2: What are the effects of knowledge-based SFT on tasks that require such knowledge and those that do not?}
 We compare the performance of pretrained LLM, LLM with single knowledge vector and LLM with combination of different knowledge vectors. The results are presented in Table \ref{tab:llm-comparison}.
% \paragraph{LLM with single knowledge vectors.}

In terms of the results using discourse relation knowledge vector, both LLaMA and Qwen show significant improvements on the Sentence Fusion task, while exhibits a slight performance decline on the GEC task. On general tasks,  LLaMA’s performance remains largely unaffected, whereas Qwen demonstrates a notable improvement.

Regarding the results using grammatical error knowledge vector, LLaMA and Qwen achieve improvements on both the Sentence Fusion and GEC tasks. However, LLaMA exhibits a slight performance decline on general tasks, whereas Qwen shows an improvement.

The results using the combined knowledge vectors of discourse relation and grammatical error exhibit a more complex pattern. Both LLaMA and Qwen achieve significant improvements on the Sentence Fusion task, reaching optimal performance, as both types of knowledge vectors contribute positively to this task. However, for the GEC task, the combined knowledge vectors lead to a slight performance decline. On general tasks, LLaMA experiences a slight decrease in performance, whereas Qwen demonstrates a significant improvement.

In summary, whether used individually or in combination, knowledge vectors can enhance the performance of LLMs on tasks that require such knowledge while not leading to a significant decline in other tasks.

The results highlight that LLM with single knowledge vector effectively balances general capabilities and knowledge-related capabilities.

\section{Related Work}
\paragraph{Learning from mistakes}
Humans can learn from mistakes to improve their capabilities and correct mistakes. Inspired by this, researchers have explored leveraging mistakes to enhance the performance of LLMs \citep{tong2024can, an2023learning, li2024turning, wang2024learning}. The LEMA (LEarning from MistAkes) method proposed by \citet{an2023learning} fine-tuning LLMs on pairs consisting of errors and their respective corrections generated by GPT-4. Similarly, \citet{tong2024can} fine-tuning LLMs on CoTErrorSet, a benchmark constructed by having the LLM prompted to correct its own errors based on the correct reference and the incorrect response generated by itself.

However, rather than fine-tuning on datasets constructed based on error responds across various tasks, we analyze the causes of errors from the perspective of knowledge deficiencies and correct errors by fine-tuning the model to learn the required knowledge from a curated knowledge dataset.

\paragraph{Self-correction}
Self-correction typically involves three stages: a LLM generates initial outputs, a feedback model generates feedback given the input and initial output and a refinement model generates a refined output considering the input, initial output and feedback. In the context of self-correction, LLMs refine their own responds based on the feedback from either themselves \citep{madaan2024self} or external tools or knowledge \citep{shinn2024reflexion,gou2023critic}.
Self-correction focus on the utilization of feedback to refine the outputs of LLM while our KSOD framework aims to improve the LLM itself from the perspective of knowledge. 

\section{Limitations}
Although our study presents a promising framework for supplementing LLMs with desired knowledge on demand, its scope is limited to knowledge of categories which can be formalized as a classification task. Future research could explore the KSOD framework to other knowledge, such as knowledge of theories and knowledge of algorithms that cannot be formalized as a classification task.

\section{Conclusion}
In this study, we introduce a novel knowledge-based SFT framework, KSOD, to supplement knowledge missing in LLMs that causes errors in outputs of LLMs. We propose a knowledge verification method and validate its effectiveness. Our framework effectively balances the LLMs' performance across both general and knowledge-related tasks. We demonstrated the effectiveness of KSOD through LLMs with both single and combination of knowledge vectors, which outperformed pre-trained LLMs on comprehensive benchmarks.

% Bibliography entries for the entire Anthology, followed by custom entries
%\bibliography{anthology,custom}
% Custom bibliography entries only
\bibliography{custom}

\begin{table*}[h]
\centering
\resizebox{0.9\textwidth}{!}{%
\begin{tabular}{>{\raggedleft\arraybackslash}p{0.2\textwidth}p{0.85\textwidth}}
\toprule[1pt]

\textbf{Task Prompt~$\to$} &
  Sentence fusion is the task of joining several independent sentences into a single coherent text. Please analyze the errors that arise in output of sentence fusion task in the given examples.
 \\ \hline
Samples~$\to$ &
% \makecell[l]{
\textbf{Example 1}:
Input: In a family tiff in 1865, his remains were exhumed and reburied in the family burial plot at Cobham Park, the family home in Warren County, Pennsylvania. Cobham's wife and in-laws sued the Cobham family for their return.
Target: In a family tiff in 1865, his remains were exhumed and reburied in the family burial plot at Cobham Park, the family home in Warren County, Pennsylvania. As a result, cobham's wife and in-laws sued the Cobham family for their return.
Output: The remains of Charles Sanders Peirce, an American philosopher and the founder of pragmatism, were exhumed and reburied in the family burial plot at Cobham Park, the family home in Warren County, Pennsylvania. Peirce's wife and in-laws sued the Cobham family for their return.
\textbf{Example 2}:
Input: He retired to Boulogne-sur-Mer, so poor that the East India Company had to vote him a pension, but later returned to England and managed eventually to pay his creditors in full. Some inheritance was left for his descendants.
Target: He retired to Boulogne-sur-Mer, so poor that the East India Company had to vote him a pension, but later returned to england and managed eventually to pay his creditors in full so that some inheritance was left for his descendants.
Output: He retired to Boulogne-sur-Mer, so poor that the East India Company had to vote him a pension, but later returned to England and managed eventually to pay his creditors in full. Some inheritance was left for his descendants.
\textbf{Example 3}:
Input: They finished third among the league's eight teams, with Gore as their starting center fielder. O'Rourke had moved to left due the departure of Slattery.
Target: They finished third among the league's eight teams, with Gore as their starting center fielder, while O'Rourke had moved to left due the departure of Slattery.
Output: They finished third among the league's eight teams, with Gore as their starting center fielder. O'Rourke had moved to left due to the departure of Slattery.
\textbf{Example 4}:
Input: The Post Office committee was a regular recipient of complaints from southern states concerning the transmission of abolitionist mailings, which were seen there as incendiary; the matter was of some controversy. Southern legislators sought to have these types of mailings banned.
Target: The Post Office committee was a regular recipient of complaints from southern states concerning the transmission of abolitionist mailings, which were seen there as incendiary; the matter was of some controversy because southern legislators sought to have these types of mailings banned.
Output: The Post Office committee, which was a regular recipient of complaints from southern states concerning the transmission of abolitionist mailings, which were seen there as incendiary, and the matter was of some controversy. Southern legislators sought to have these types of mailings banned.
 \\ \hline
\texttt{Chain-of-thought prompt}~$\to$ &
Firstly, provide a step-by-step analysis for the common characteristics of the errors from all examples. 
Next, identify the potential knowledge lacking in LLM that may have led to these errors.
 \\ 
  
\bottomrule[1pt]
\end{tabular}
}
\caption{Complement prompt of knowledge identification.}
\label{tab:KI_prompt}
\end{table*}

\newpage
\appendix
% \section{Knowledge Dimension of Bloom's Taxonomy}
% \label{A:knowledge_dimension}

% \section{Tasks and Datasets}
% \label{A:identification}
\section{Prompts for Knowledge Identification}
\label{app:identification}
The templates used for knowledge identification in the knowledge identification stage are presented in Table \ref{tab:KI_prompt}.
\section{Knowledge verification Settings}
\label{app:verification}
\subsection{Details of training knowledge module.}
The learning rate for both classification linear layer and LoRA is set to 5e-5. For the rank of LoRA, we select the value that yields the best model performance from {8, 16, 32, 64} for the final experiments. The target module of LoRA is the output matrix in the last self-attention layer. 

\subsection{Evaluation on benchmarks.}
\paragraph{Metrics.}
For four general benchmarks, we use the default settings in LLMBox. For Disco Fuse task, we use SARI \citep{xu2016optimizing} to evaluate LLMs' performance. For GEC, we calculate F0.5 of $M^2$ score \citep{dahlmeier2012better} to evaluate LLMs' performance.
\paragraph{Inference settings.}
For four general benchmarks, we use the default settings in LLMBox.
For Disco Fuse task, we set temperature to 1.0. top-P to 0.9.
For GEC task, we set temperature to 0.5, top-K to 50.

\end{document}